\documentclass[11pt]{article}
\pdfoutput=1
\usepackage{acl2015}
\usepackage{times}
\usepackage{url}
\usepackage{latexsym}

\usepackage[fleqn]{amsmath}
\usepackage{bm}
\usepackage{multirow}
\usepackage{graphicx}
\usepackage{caption}
\usepackage{subcaption}
\usepackage{enumitem}
\usepackage{color}
\usepackage[usenames,dvipsnames,svgnames,table]{xcolor}
\usepackage{verbatim}
\usepackage{fmtcount} 
\usepackage[noend]{algorithmic}
\usepackage{algorithm}
\renewcommand{\algorithmicforall}{for each}
\renewcommand{\algorithmicwhile}{while}
\renewcommand\algorithmicdo{}

\newcommand{\ignore}[1]{}

\title{A Bayesian Model of Multilingual \\ Unsupervised Semantic Role Induction}

\author{Nikhil Garg \\
  University of Geneva \\
  Switzerland \\
  {\tt nikgarg@gmail.com} \\\And
  James Henderson \\
  University of Geneva \\
  Switzerland \\
  {\tt james.henderson@unige.ch} \\}

\date{}

\begin{document}
\maketitle
\begin{abstract}
We propose a Bayesian model of unsupervised semantic role induction in
multiple languages, and use it to explore the usefulness of parallel corpora
for this task.  Our joint Bayesian model consists of individual models for
each language plus additional latent variables that capture alignments between
roles across languages.  Because it is a generative Bayesian model, we can do
evaluations in a variety of scenarios just by varying the inference procedure,
without changing the model, thereby comparing the scenarios directly.  We compare using only monolingual data, using a parallel
corpus, using a parallel corpus with annotations in the other language, and
using small amounts of annotation in the target language.  We find that the
biggest impact of adding a parallel corpus to training is actually the
increase in mono-lingual data, with the alignments to another language resulting in small improvements, 
even with labeled data for the other language. 
\end{abstract}

\section{Introduction}
\label{sec:Introduction}
Semantic Role Labeling (SRL) has emerged as an important task in Natural Language Processing (NLP) due to its applicability in information extraction, question answering, and other NLP tasks. SRL is the problem of finding predicate-argument structure in a sentence, as illustrated below:
\vspace{-1ex}
\[[_{A0} \text{ Mike }] \text{ has } [_{PRED} \text{ written }] [_{A1} \text{ a book }] \quad (S1)
\vspace{-1ex}
\]
Here, the predicate WRITE has two arguments: `Mike' as \emph{A0} or \emph{the writer}, and `a book' as \emph{A1} or \emph{the thing written}. The labels \emph{A0} and \emph{A1} correspond to the PropBank annotations \cite{palmer2005proposition}.

As the need for SRL arises in different domains and languages, the existing
manually annotated corpora become insufficient to build supervised
systems. This has motivated work on unsupervised SRL \cite{lang2011unsupervisedEMNLP,titov2012bayesian,garg2012unsupervised}. Previous work has indicated that unsupervised systems could benefit from the word alignment information in parallel text in two or more languages \cite{naseem2009multilingual,snyder2009unsupervised,titovcrosslingual}. For example, consider the German translation of sentence $S1$:
\vspace{-1ex}
\[[_{A0}\text{Mike}] \text{ hat } [_{A1}\text{ein Buch}] [_{PRED}\text{geschrieben}] \quad (S2)
\vspace{-1ex}
\]
If sentences $S1$ and $S2$ have the word alignments: Mike-Mike, written-geschrieben, and book-Buch, the system might be able to predict \emph{A1} for Buch, even if there is insufficient information in the monolingual German data to learn this assignment. Thus, in languages where the resources are sparse or not good enough, or the distributions are not informative, SRL systems could be made more accurate by using parallel data with resource rich or more amenable languages. 

In this paper, we propose a joint Bayesian model for unsupervised semantic
role induction in multiple languages. The model consists of individual
Bayesian models for each language \cite{garg2012unsupervised}, and
\emph{crosslingual latent variables} to incorporate soft role agreement
between aligned constituents. This latent variable approach has been
demonstrated to increase the performance in a multilingual unsupervised
part-of-speech tagging model based on HMMs \cite{naseem2009multilingual}. We
investigate the application of this approach to unsupervised SRL, presenting the performance improvements obtained in different settings involving labeled and unlabeled data, and analyzing the annotation effort required to obtain similar gains using labeled data.

We begin by briefly describing the unsupervised SRL pipeline and the monolingual semantic role induction model we use, and then describe our multilingual model.

\section{Unsupervised SRL Pipeline}
\label{sec:USRLPipeline}
As established in previous work \cite{gildea2002automatic,pradhan2005support},
we use a standard unsupervised SRL setup, consisting of the following steps:
\begin{description}[leftmargin=*] \itemsep0em
\item[1. Syntactic Parsing] Off-the-shelf parsers can be used to syntactically parse a given sentence. We use a dependency parse because of its simplicity and easier comparison with the previous work in unsupervised SRL.
\item[2. Predicate Identification] We select all the non-auxiliary verbs in a sentence as predicates.
\item[3. Argument Identification] For a given predicate, this step classifies each constituent of the parse tree as a semantic argument or a non-argument. Heuristics based on syntactic features such as the dependency relation of a constituent to its head, path from the constituent to the predicate, etc.\ have been used in unsupervised SRL.
\item[4. Argument Classification] Without access to semantic role labels, unsupervised SRL systems cast the problem as a clustering problem. Arguments of a predicate in all the sentences are divided into clusters such that each cluster corresponds to a single semantic role. The better this clustering is, the easier it becomes for a human to give it an actual semantic role label like \emph{A0}, \emph{A1}, etc. Our model assigns a role variable to every identified argument. This variable can take any value from \(1\) to \(N\), where \(N\) is the number of semantic roles that we want to induce.
\end{description}
The task we model, unsupervised semantic role induction, is the step~4 of this
pipeline. 

\section{Monolingual Model}
\label{sec:MonolingualModel}
We use the Bayesian model of \newcite{garg2012unsupervised} as our base
monolingual model.  The semantic roles are predicate-specific. To model
the role ordering and repetition preferences, the role inventory for each
predicate is divided into \emph{Primary} and \emph{Secondary} roles as
follows:
\vspace{-1mm}
\begin{description}[leftmargin=*] \itemsep0em
\item[Primary Role (PR)] Let there be a total of \(N\) roles (or clusters) for each predicate. Assign \(K\) of them as PRs {\footnotesize\(\{P_1, P_2,...,P_K\}\)}. Further, create 3 additional PRs: {\footnotesize\(START\)} denoting the start of the role sequence, {\footnotesize\(END\)} denoting its end, and {\footnotesize\(PRED\)} denoting the predicate. These \((K+3)\) PRs are not allowed to repeat in a frame and their ordering defines the global role ordering.
\item[Secondary Role (SR)] The rest of the \((N-K)\) roles are called SRs {\footnotesize\(\{S_1,S_2,...,S_{N-K}\}\)}. Unlike PRs, they are not constrained to occur only once and only their ordering w.r.t.\ PRs is used in the probability model.
\end{description}

For example, the complete role sequence in a frame could be:
{\footnotesize\((\)\(START\), \(P_3\), \(S_1\), \(S_1\), \(PRED\), \(P_2\),
  \(S_5\), \(END\)\()\)}. The \emph{ordering} is defined as the sequence of
PRs, {\footnotesize\((\)\(START\), \(P_3\), \(PRED\), \(P_2\),
  \(END\)\()\)}. Each pair of consecutive PRs in an ordering is called an
\emph{interval}. Thus, {\footnotesize\((P_3,PRED)\)} is an interval that
contains two SRs, \(S_1\) and \(S_1\). An interval could also be empty, for
instance {\footnotesize\((START,P_3)\)} contains no SRs.  When we evaluate,
these roles get mapped to gold roles. For instance, the PR
{\footnotesize\(P_2\)} could get mapped to a core role like
{\footnotesize\(A0\), \(A1\),} etc.\ or to a modifier role like
{\footnotesize\(AM-TMP\), \(AM-MOD\),} etc. \newcite{garg2012unsupervised}
reported that, in practice, PRs mostly get mapped to core roles and SRs to
modifier roles, which conforms to the linguistic motivations for this
distinction.

\begin{figure*}[!ht]

  \begin{subfigure}{\textwidth}
    \begin{flalign}
      \label{eq:joint_Model1}
      &P(\mathbf{r}, \mathbf{f} | p, vc) =  \underbrace{P(o|p,vc)}_{o = ordering(\mathbf{r})} \qquad \underbrace{\prod_{r_i \in \mathbf{r} \cap PR}{P(f_i|r_i,p)}}_\text{Primary Roles} \qquad \underbrace{\prod_{I \in o} P(\mathbf{r}(I),\mathbf{f}(I)|I,p)}_\text{Intervals}& \\
      \label{eq:prob_interval}
      &\mbox{\scriptsize{where}} \; P(\mathbf{r}(I),\mathbf{f}(I) | I, p) = \prod_{r_i \in \mathbf{r}(I)} {\underbrace{P(\neg stop|I,p,adj)}_\text{generate indicator} \underbrace{P(r_i|I,p)}_\text{generate SR} \underbrace{P(f_i|r_i,p)}_\text{generate features}} \; \underbrace{P(stop|I,p,adj)}_\text{end of the interval}& \\
      \label{eq:prob_constituent}
      &\mbox{\scriptsize{and}} \quad P(f_i|r_i,p) = \prod_{t=1}^{T} {P(f_{i,t}|r_i,p)}& \\
      \label{eq:marginal_Model1}
      &P(\mathbf{f} | p, vc) =  \sum_{\mathbf{r}} {P(\mathbf{r}, \mathbf{f} | p, vc)}&
    \end{flalign}
    \vspace{-5mm}
    \caption{\label{fig:Model1_eq} Probability equations for the monolingual model. Bold-faced variables denote a sequence of values. \(\mathbf{r}\) denotes the complete sequence of roles, and \(\mathbf{f}\) denotes the complete sequence of features. \(p\) and \(vc\) denote the predicate and its voice respectively. \(o\) denotes the ordering of PRs in the sequence \(\mathbf{r}\) and \(ordering(\mathbf{r})\) is a function for computing this ordering. \(r_i\) and \(f_i\) denote the role and features at position \(i\) respectively, and \(r(I)\) and \(f(I)\) respectively denote the SR sequence and feature sequence in interval \(I\). \(f_{i,t}\) denotes the value of feature \(t\) at position \(i\). \(adj=0\) for generating the first SR, and \(1\) for a subsequent one. Equation \ref{eq:joint_Model1} gives the joint probability of the model and equation \ref{eq:marginal_Model1} gives the marginal probability of the observed features.}
  \end{subfigure}
 
  \begin{subfigure}{\textwidth}
    \begin{align}
      \label{eq:joint_bilingual}
      P(\mathbf{r}^{l1}, \mathbf{f}^{l1}, \mathbf{r}^{l2}, \mathbf{f}^{l2}, \mathbf{z} | p^{l1}, vc^{l1}, p^{l2}, vc^{l2}) &= P(\mathbf{z}) \prod_{l \in \{l1,l2\}} {P(\mathbf{r}^{l}, \mathbf{f}^{l} | \mathbf{z}, p^{l}, vc^{l})}\\
      \label{eq:joint_bilingual_approx}
      &\approx P(\mathbf{z}) \prod_{l \in \{l1,l2\}} {P(\mathbf{r}^{l}, \mathbf{f}^{l} | p^{l}, vc^{l}) \prod_{i,k: z_k \rightarrow r^{l}_i} {P(r^{l}_i | z_k)}}
    \end{align}
    \vspace{-5mm}
    \caption{\label{fig:bilingual_eq} Probability equations for the multilingual model. The superscript \(l\) denotes the variable for language \(l\). \(\mathbf{z}\) denotes the common crosslingual latent variables for both languages. \(z_k \rightarrow r^{l}_i\) denotes that the argument at position \(i\) in language \(l\) is connected to the crosslingual latent variable \#\(k\).}
  \end{subfigure}
  \vspace{-3mm}
  \caption{Probability equations for the (a) monolingual and (b) multilingual model.}
  \label{fig:model_eq}
  \vspace{-4mm}
\end{figure*}

Figure~\ref{fig:bilingual_model} illustrates two copies of the monolingual
model, on either side of the crosslingual latent variables.  The generative
process is as follows:
\vspace{-2mm}
\begin{description}[leftmargin=*] \itemsep0em
  \item[1. Predicate, Voice] The predicate \(p\) and its voice \(vc\) are treated as top-level visible variables.
  \item[2. Ordering (Generate PRs)] Select an ordered set of PRs from a multinomial distribution.\\
    \(o \sim Multinomial(\theta^{\scriptscriptstyle order}_{p,vc})\)
  \item[3. Generate SRs] For each interval in the ordering \(o\), a sequence of SRs is generated as:\\
    \begin{tabular}{|l}
      for each interval \(I \in o\):\\
      \hspace*{1mm} draw an indicator \(s \sim Binomial(\theta^{\scriptscriptstyle STOP}_{p,I,0})\)\\
      \hspace*{1mm} while \(s \neq {\scriptstyle STOP}\):\\
      \hspace*{3mm} choose a SR \(r \sim Multinomial(\theta^{\scriptscriptstyle SR}_{p,I})\)\\
      \hspace*{3mm} draw an indicator \(s \sim Binomial(\theta^{\scriptscriptstyle STOP}_{p,I,1})\)
    \end{tabular}
  \item[4. Generate Features] For each PR and SR, the features for that constituent are generated independently. To keep the model simple and comparable to previous unsupervised work, we only use three features: (i) dependency relation of the argument to its head, (ii) head word of the argument, and (iii) POS tag of the head word:\\
    \begin{tabular}{|l}
      for each generated role \(r\):\\
      \hspace*{0.5mm} for each feature type \(f\):\\
      \hspace*{2mm} choose a value \(v_f \sim Multinomial(\theta^{\scriptscriptstyle F}_{p,r,f})\)
    \end{tabular}
\end{description}

All the multinomial and binomial distributions have symmetric Dirichlet and beta priors respectively. Figure \ref{fig:Model1_eq} gives the probability equations for the monolingual model. This formulation models the global role ordering and repetition preferences using PRs, and limited context for SRs using intervals. Ordering and repetition information was found to be helpful in supervised SRL as well \cite{punyakanok2004semantic,pradhan2005support,toutanova2008global}. More details, including the motivations behind this model, are in \cite{garg2012unsupervised}.

\section{Multilingual Model}
\label{sec:MultilingualModel}
\begin{figure*}
  \centering
  \includegraphics[width=0.9\textwidth]{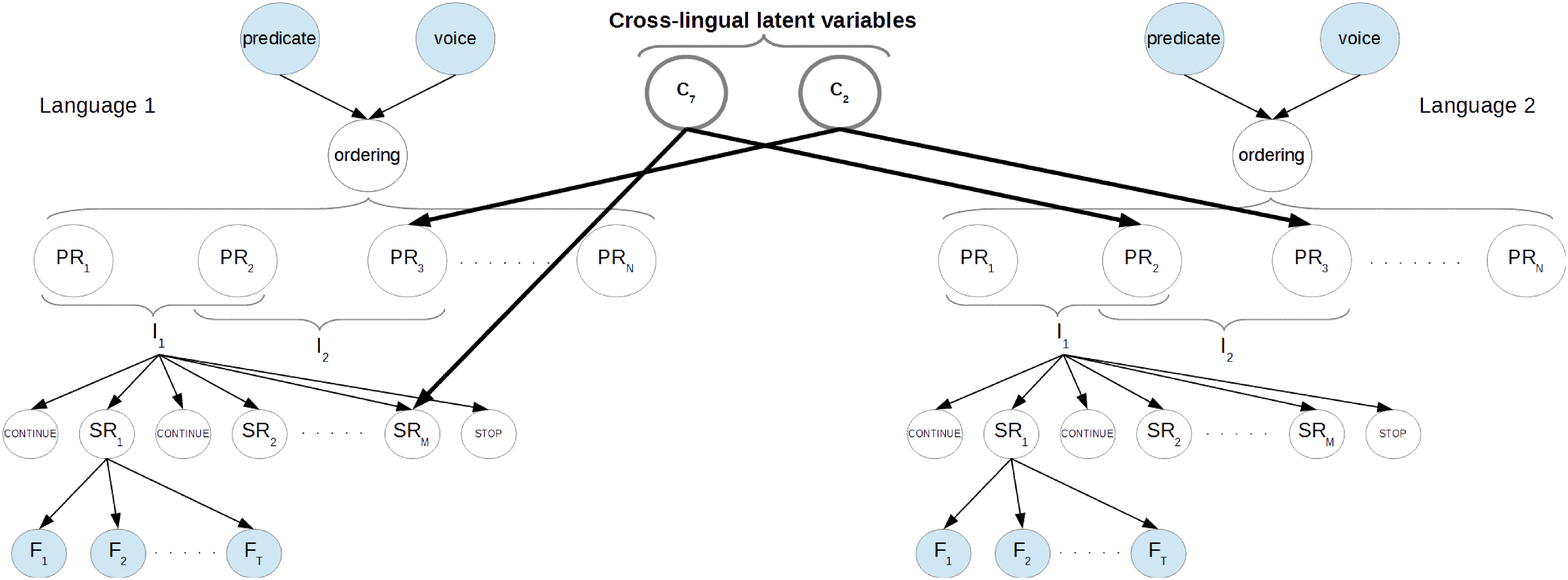}
  \caption{Multilingual model. The CLVs and their associated parameters are drawn in bold. {\footnotesize\(PR_3\)} in language 1 is aligned to {\footnotesize\(PR_3\)} in language 2 with the corresponding CLV taking the value \(c_2\), and {\footnotesize\(SR_M\)} is aligned to {\footnotesize\(PR_2\)} with the CLV taking the value \(c_7\).}
  \label{fig:bilingual_model}
\vspace{-4mm}
\end{figure*}

The multilingual model uses word alignments between sentences in a parallel
corpus to exploit role correspondences across languages. We make copies of the
monolingual model for each language and add additional \emph{crosslingual
  latent variables} (CLVs) to couple the monolingual models, capturing crosslingual
semantic role patterns. Concretely, when training on parallel sentences,
whenever the head words of the arguments are aligned, we add a CLV as a parent
of the two corresponding role variables. Figure \ref{fig:bilingual_model}
illustrates this model. The generative process, as explained below, remains
the same as the monolingual model for the most part, with the exception of
aligned roles which are now generated by both the monolingual process as well
as the CLV.
\begin{description}[leftmargin=*] \itemsep0em
  \item[1. Monolingual Data] Given a parallel frame with the predicate pair \(p1,p2\), generate two separate monolingual frames as in section \ref{sec:MonolingualModel}.
  \item[2. Aligned Arguments] For each aligned argument, first generate a
    crosslingual latent variable from a Chinese Restaurant Process (CRP). Then generate the two aligned roles:\\
    \begin{tabular}{|l}
      for aligned arguments \(i,j\):\\
      \hspace*{1mm} draw a crosslingual latent variable:\\
      \hspace*{5mm} \(z \sim CRP(\alpha^{\scriptscriptstyle CRP}_{p1,p2})\)\\
      \hspace*{1mm} draw role for language \(l1\):\\
      \hspace*{5mm} \(r_i \sim Multinomial(\theta^{\scriptscriptstyle align}_{p1,p2,z,l1})\)\\
      \hspace*{1mm} draw role for language \(l2\):\\
      \hspace*{5mm} \(r_j \sim Multinomial(\theta^{\scriptscriptstyle align}_{p1,p2,z,l2})\) 
    \end{tabular}
\end{description}

Every predicate-tuple has its own inventory of CLVs specific to that tuple. Each CLV \(z\) is a multi-valued variable where each value defines a distribution over role labels for each language (denoted by \(\theta^{\scriptscriptstyle align}_{p1,p2,z,l}\) above). 
These distributions over labels are trained to be peaky, so that each value \(c\) for a CLV represents a correlation between the labels that \(c\) predicts in the two languages.  
For example, a  value \(c\) for the CLV \(z\) might give high probabilities to {\footnotesize\(S_3\)} and {\footnotesize\(S_8\)} in language 1, and to {\footnotesize\(S_1\)} in language 2.  If \(c\) is the only value for \(z\) that gives high probability to {\footnotesize\(S_3\)} in language 1, and the monolingual model in language 1 decides to assign {\footnotesize\(S_3\)} to the role for \(z\), then \(z\) will predict {\footnotesize\(S_1\)} in language 2, with high probability.
We generate the CLVs via a Chinese Restaurant Process \cite{pitman2002combinatorial}, a non-parametric Bayesian model, which allows us to induce the number of CLVs for every predicate-tuple from the data. We continue to train on the non-parallel sentences using the respective monolingual models.

The multilingual model is deficient, since the aligned roles are being generated twice. Ideally, we would like to add the CLV as additional conditioning variables in the monolingual models. The new joint probability can be written as equation \ref{eq:joint_bilingual} (Figure \ref{fig:bilingual_eq}), which can be further decomposed following the decomposition of the monolingual model in Figure \ref{fig:Model1_eq}. However, having this additional conditioning variable breaks the Dirichlet-multinomial conjugacy, which makes it intractable to marginalize out the parameters during inference. Hence, we use an approximation where we treat each of the aligned roles as being generated twice, once by the monolingual model and once by the corresponding CLV (equation \ref{eq:joint_bilingual_approx}).

This is the first work to incorporate the coupling of aligned arguments directly in a Bayesian SRL model.  This makes it easier to see how to extend this model in a principled way to incorporate additional sources of information.  First, the model scales gracefully to more than two languages. If there are a total of \(n\) languages, and there is an aligned argument in \(m\) of them, the multilingual latent variable is connected to only those \(m\) aligned arguments. 

Second, having one joint Bayesian model allows us to use the same model in various semi-supervised learning settings, just by fixing the annotated variables during training. Section~\ref{ssec:projection} evaluates a setting where we have some labeled data in one language (called source), while no labeled data in the second language (called target). Note that this is different from a classic annotation projection setting (e.g.\ \cite{pado2009cross}), where the role labels are mapped from source constituents to aligned target constituents.

\section{Inference and Training}
\label{sec:InferenceAndTraining}
The inference problem consists of predicting the role labels and CLVs (the hidden variables) given the predicate, its voice, and syntactic features of all the identified arguments (the visible variables). We use a collapsed Gibbs-sampling based approach to generate samples for the hidden variables (model parameters are integrated out). The sample counts and the priors are then used to calculate the MAP estimate of the model parameters.

For the monolingual model, the role at a given position is sampled as:
\vspace{-2mm}
\begin{flalign}
  &\scriptstyle P(r_i | \mathbf{r}_{-i}, \mathbf{f}, p, vc, D^-) \propto P(r_i, \mathbf{r}_{-i}, \mathbf{f} | p, vc, D^-)& \nonumber \\
  &\scriptstyle = \int P(r_i, \mathbf{r}_{-i}, \mathbf{f} | \bm{\theta}, p, vc) P(\bm{\theta} | D^-) \mathrm{d}\bm{\theta}& \nonumber
\end{flalign}
where the subscript \(-i\) refers to all the variables except at position \(i\), \(D^-\) refers to the variables in all the training instances except the current one, and \(\bm{\theta}\) refers to all the model parameters. The above integral has a closed form solution due to Dirichlet-multinomial conjugacy.

For sampling roles in the multilingual model, we also need to consider the probabilities of roles being generated by the CLVs:
\vspace{-2mm}
\begin{flalign}
  &\scriptstyle P(r_i | \mathbf{r}_{-i}, \mathbf{f}, p, vc, \mathbf{z}, D^-) \propto P(r_i, \mathbf{r}_{-i}, \mathbf{f} | \mathbf{z}, p, vc, D^-)& \nonumber \\
  &\scriptstyle = \int P(r_i, \mathbf{r}_{-i}, \mathbf{f} | \bm{\theta}, \mathbf{z}, p, vc) P(\bm{\theta} | D^{-}) \mathrm{d}\bm{\theta}& \nonumber \\
  &\scriptstyle = \int P(r_i, \mathbf{r}_{-i}, \mathbf{f} | \bm{\theta}, p, vc) (\underset{j,k: z_k \rightarrow r_j \hfill}{\prod {P(r_j | \bm{\theta}, z_k)}}) P(\bm{\theta} | D^-) \mathrm{d}\bm{\theta}& \nonumber
\end{flalign}

For sampling CLVs, we need to consider three factors: two corresponding to probabilities of generating the aligned roles, and the third one corresponding to selecting the CLV according to CRP.
\vspace{-2mm}
\begin{flalign}
  &\scriptstyle P(z_k | r^{l1}_i, r^{l2}_j, D^{-,k}) \propto P(r^{l1}_i | z_k, D^{-,k}) P(r^{l2}_j | z_k, D^{-,k}) P(z_k | D^{-,k})& \nonumber
\end{flalign}
where the aligned roles \(r^{l1}_i\) and \(r^{l2}_j\) are connected to \(z_k\), and \(D^{-,k}\) refers to all the variables except \(z_k\), \(r^{l1}_i\), and \(r^{l2}_j\).

We use the trained parameters to parse the monolingual data using the monolingual model. The crosslingual parameters are ignored even if they were used during training. Thus, the information coming from the CLVs acts as a regularizer for the monolingual models.

\section{Experiments}
\label{sec:Experiments}

\subsection{Evaluation}
\label{ssec:Evaluation}
Following the setting of \newcite{titovcrosslingual}, we evaluate only on the arguments that were correctly identified, as the incorrectly identified arguments do not have any gold semantic labels. Evaluation is done using the metric proposed by \newcite{lang2011unsupervised}, which has 3 components: (i) \textbf{Purity (PU)} measures how well an induced cluster corresponds to a single gold role, (ii) \textbf{Collocation (CO)} measures how well a gold role corresponds to a single induced cluster, and (iii) \textbf{F1} is the harmonic mean of PU and CO. For each predicate, let \(N\) denote the total number of argument instances, \(C_i\) the instances in the induced cluster \(i\), and \(G_j\) the instances having label \(j\) in gold annotations.
\( \scriptstyle PU = \frac{1}{N} \sum_i max_j |C_i \cap G_j| \) , \( \scriptstyle CO = \frac{1}{N} \sum_j max_i |C_i \cap G_j| \) , and \( \scriptstyle F1 = \frac{2 \cdot PU \cdot CO}{PU + CO} \). 
The score for each predicate is weighted by the number of its argument instances, and a weighted average is computed over all the predicates.

\subsection{Baseline}
\label{ssec:Baseline}
We use the same baseline as used by \newcite{lang2011unsupervised} which has been shown to be difficult to outperform. This baseline assigns a semantic role to a constituent based on its syntactic function, i.e.\ the dependency relation to its head. If there is a total of \(N\) clusters, \((N-1)\) most frequent syntactic functions get a cluster each, and the rest are assigned to the \(N\)th cluster.

\subsection{Closest Previous Work}
\label{ssec:PreviousWork}
This work is closely related to the cross-lingual unsupervised SRL work of \newcite{titovcrosslingual}. Their model has separate monolingual models for each language and an extra penalty term which tries to maximize \(P(r^{l2}|r^{l1})\) and \(P(r^{l1}|r^{l2})\) i.e.\ for all the aligned arguments with role label \(r^{l1}\) in language 1, it tries to find a role label \(r^{l2}\) in language 2 such that the given proportion is maximized and vice verse. However, there is no efficient way to optimize the objective with this penalty term and the authors used an inference method similar to annotation projection. Further, the method does not scale naturally to more than two languages. Their algorithm first does monolingual inference in one language ignoring the penalty and then does the inference in the second language taking into account the penalty term. In contrast, our model adds the latent variables as a part of the model itself, and not an external penalty, which enables us to use the standard Bayesian learning methods such as sampling.

The monolingual model we use \cite{garg2012unsupervised} also has two main advantages over \newcite{titovcrosslingual}. First, the former incorporates a global role ordering probability that is missing in the latter. Secondly, the latter defines \emph{argument-keys} as a tuple of four syntactic features and all the arguments having the same argument-keys are assigned the same role. This kind of hard clustering is avoided in the former model where two constituents having the same set of features might get assigned different roles if they appear in different contexts.  

\subsection{Data}
\label{ssec:Data}
Following \newcite{titovcrosslingual}, we run our experiments on the English (EN) and German (DE) sections of the CoNLL 2009 corpus \cite{hajic2009conll}, and EN-DE section of the Europarl corpus \cite{koehn2005europarl}. We get about 40k EN and 36k DE sentences from the CoNLL 2009 training set, and about 1.5M parallel EN-DE sentences from Europarl. For appropriate comparison, we keep the same setting as in \cite{titovcrosslingual} for automatic parses and argument identification, which we briefly describe here. The EN sentences are parsed syntactically using MaltParser \cite{nivre2007maltparser} and DE using LTH parser \cite{johansson2008dependency}. All the non-auxiliary verbs are selected as predicates. In CoNLL data, this gives us about 3k EN and 500 DE predicates. The total number of predicate instances are 3.4M in EN (89k CoNLL + 3.3M Europarl) and 2.62M in DE (17k CoNLL + 2.6M Europarl). The arguments for EN are identified using the heuristics proposed by \newcite{lang2011unsupervised}. However, we get an F1 score of 85.1\% for argument identification on CoNLL 2009 EN data as opposed to 80.7\% reported by \newcite{titovcrosslingual}. This could be due to implementation differences, which unfortunately makes our EN results incomparable. For DE, the arguments are identified using the LTH system \cite{johansson2008dependency}, which gives an F1 score of 86.5\% on the CoNLL 2009 DE data. The word alignments for the EN-DE parallel Europarl corpus are computed using GIZA++ \cite{och2003systematic}. For high-precision, only the intersecting alignments in the two directions are kept. We define two semantic arguments as aligned if their head-words are aligned. In total we get 9.3M arguments for EN (240k CoNLL + 9.1M Europarl) and 4.43M for DE (32k CoNLL + 4.4M Europarl). Out of these, 0.76M arguments are aligned.

\subsection{Main Results}
\label{ssec:Results}

\begin{table*}
  \centering
  \begin{tabular}{| l | l | l | l | c  c  c | c c c |}
    \hline
    & \multicolumn{2}{| c |}{\textbf{Dataset}} & \textbf{Model} & \multicolumn{3}{|c|}{\textbf{English}} & \multicolumn{3}{|c|}{\textbf{German}} \\
    & \textbf{Training} & \textbf{Testing} & & \textbf{PU} & \textbf{CO} & \textbf{F1} & \textbf{PU} & \textbf{CO} & \textbf{F1} \\
    \hline
    0 & CoNLL & CoNLL & Baseline & 78.23 & 79.46 & 78.84 & 83.09 & 79.32 & 81.16 \\
    \hline
    1 & CoNLL & CoNLL & Monolingual & 76.29 & 83.13 & 79.56* & 82.54 & 81.94 & 82.24* \\ 
    2 & CoNLL+EP & CoNLL & Monolingual & 76.11 & 83.33 & 79.56 & 83.77 & 81.65 & 82.70* \\ 
    3 & 2$\times$(CoNLL+EP) & CoNLL & Multilingual & 76.23 & 83.25 & \textbf{79.59} & 83.81 & 81.79 & \textbf{82.79} \\ 
    \hline
    4 & EP & CoNLL & Monolingual & 73.26 & 80.60 & 76.76 & 83.72 & 81.28 & 82.48 \\
    5 & 2$\times$EP & CoNLL & Multilingual & 73.07 & 81.24 & \textbf{76.94} & 83.59 & 81.50 & \textbf{82.54} \\ 
    \hline
  \end{tabular}
  \vspace{-2mm}
  \caption{Main Results. A * denotes a significant improvement in the F1 score over the the previous line. We compute the significance using stratified shuffling and consider it significant if the p-value \(< 0.05\).}
  \label{tab:results}
\vspace{-1mm}
\end{table*}

Since the CoNLL annotations have 21 semantic roles in total, we use 21 roles in our model as well as the baseline. Following \newcite{garg2012unsupervised}, we set the number of PRs to 2 (excluding {\footnotesize\(START\)}, {\footnotesize\(END\)} and {\footnotesize\(PRED\)}), and SRs to 21-2=19. Table \ref{tab:results} shows the results.

In the first setting (Line 1), we train and test the monolingual model on the CoNLL data.  We observe significant improvements in F1 score over the Baseline (Line 0) in both languages. Using the CoNLL 2009 dataset alone, \newcite{titovcrosslingual} report an F1 score of 80.9\% (PU=86.8\%, CO=75.7\%) for German.  Thus, our monolingual model outperforms their monolingual model in German.  For English, they report an F1 score of 83.6\% (PU=87.5\%, CO=80.1\%), 
but note that our English results are not directly comparable to theirs due to differences argument identification, as discussed in section \ref{ssec:Data}. As their argument identification score is lower, perhaps their system is discarding ``difficult'' arguments which leads to a higher clustering score.

In the second setting (Line 2), we use the additional monolingual Europarl (EP) data for training. We get equivalent results in English and a significant improvement in German compared to our previous setting (Line 1). The German dataset in CoNLL is quite small and benefits from the additional EP training data. In contrast, the English model is already quite good due to a relatively big dataset from CoNLL, and good accuracy syntactic parsers. Unfortunately, \newcite{titovcrosslingual} do not report results with this setting.

The third setting (Line 3) gives the results of our multilingual model, which adds the word alignments in the EP data. Comparing with Line 2, we get non-significant improvements in both languages.  \newcite{titovcrosslingual} obtain an F1 score of 82.7\% (PU=85.0\%, CO=80.6\%) for German, and 83.7\% (PU=86.8\%, CO=80.7\%) for English. Thus, for German, our multilingual Bayesian model is able to capture the cross-lingual patterns at least as well as the external penalty term in \cite{titovcrosslingual}. We cannot compare the English results unfortunately due to differences in argument identification.

We also compared monolingual and bilingual training data using a setting that emulates the standard supervised setup of separate training and test data sets. We train only on the EP dataset and test on the CoNLL dataset. Lines 4 and 5 of Table~\ref{tab:results} give the results. The multilingual model obtains small improvements in both languages, which confirms the results from the standard unsupervised setup, comparing lines 2 to 3.

\begin{table*}
  \centering
  \begin{tabular}{| l | l | l | c  c  c | c c c |}
    \hline
    & \textbf{Source} & \textbf{Target} & \multicolumn{3}{|c|}{\textbf{English}} & \multicolumn{3}{|c|}{\textbf{German}} \\
    & & & \textbf{PU} & \textbf{CO} & \textbf{F1} & \textbf{PU} & \textbf{CO} & \textbf{F1} \\
    \hline
    1 & \multicolumn{2}{| c | }{Multilingual Model} & 76.23 & 83.25 & 79.59 & 83.81 & 81.79 & 82.79 \\
    \hline
    2 & English & German & \multicolumn{3}{| c |}{NA} & 83.83 & 81.83 & \textbf{82.82} \\ 
    3 & German & English & 76.26 & 83.37 & \textbf{79.66} & \multicolumn{3}{| c |}{NA} \\ 
    \hline
  \end{tabular}
  \vspace{-2mm}
  \caption{Results for the Multilingual Model with using labeled data for the source language.}
  \label{tab:labeled_source}
  \vspace{-2mm}
\end{table*}

These results indicate that little information can be learned about semantic roles from this parallel data setup.  
One possible explanation for this result is that the setup itself is inadequate.  Given the definition of aligned arguments, only 8\% of English arguments and 17\% of German arguments are aligned.  This plus our experiments suggest that improving the alignment model is a necessary step to making effective use of parallel data in multilingual SRI, for example by joint modeling with SRI.
We leave this exploration to future work.

\subsection{Multilingual Training with Labeled Data for One Language}
\label{ssec:projection}

Another motivation for jointly modeling SRL in multiple languages is the transfer of information from a resource rich language to a resource poor language.
We evaluated our model in a very general annotation transfer scenario, where we have a small labeled dataset for one language (source), and a large parallel unlabeled dataset for the source and another (target) language. We investigate whether this setting improves the parameter estimates for the target language. To this end, we clamp the role annotations of the source language in the CoNLL dataset using a predefined mapping\footnote{\(A0\) was mapped to the primary role \(P_1\), \(A1\) to \(P_2\), and the rest were mapped to the secondary roles \((S_1,... ,S_{19})\) in the order of their decreasing frequency.}, and do not sample them during training.
This data gives us good parameters for the source language, which are used to sample the roles of the source language in the unlabeled Europarl data. The CLVs aim to capture this improvement and thereby improve sampling and parameter estimates for the target language. Table \ref{tab:labeled_source} shows the results of this experiment. We obtain small improvements in the target languages. 
As in the unsupervised setting, the small percentage of aligned roles probably limits the impact of the cross-lingual information.

\subsection{Labeled Data in Monolingual Model}
\label{ssec:labeled_data}
We explored the improvement in the monolingual model in a semi-supervised setting. To this end, we randomly selected \(S\%\) of the sentences in the CoNLL dataset as ``supervised sentences'' and the rest \((100-S)\%\) were kept unsupervised. Next, we clamped the role labels of the supervised sentences using the predefined mapping from Section~\ref{ssec:projection}. Sampling was done on the unsupervised sentences as usual. We then measured the clustering performance using the trained parameters.\footnote{To account for the randomness in selecting the supervised sentences, the experiment was repeated 10 times and average of the performance numbers was taken.}

To access the contribution of partial supervision better, we constructed a ``supervised baseline'' as follows. For predicates seen in the supervised sentences, a MAP estimate of the parameters was calculated using the predefined mapping. For the unseen predicates, the standard baseline was used.
\begin{figure}[tb]
  \centering
  \begin{subfigure}[b]{0.4\textwidth}
    \includegraphics[totalheight=0.3\textheight, angle=-90]{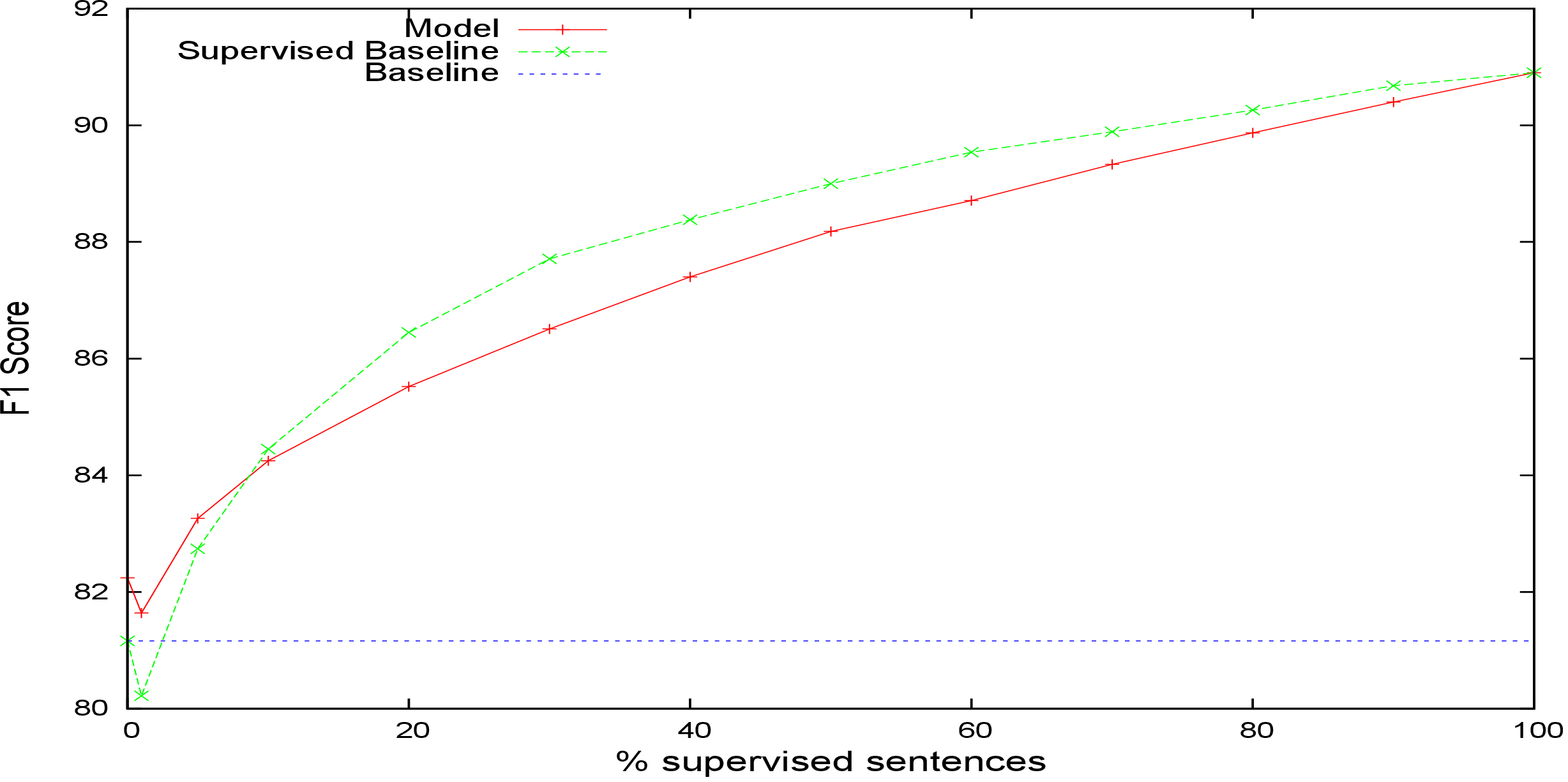}
    \vspace{-2mm}
    \caption{German}
    \label{fig:monolingual_semi_supervised_DE}
  \end{subfigure}
  \begin{subfigure}[b]{0.4\textwidth}
    \includegraphics[totalheight=0.3\textheight, angle=-90]{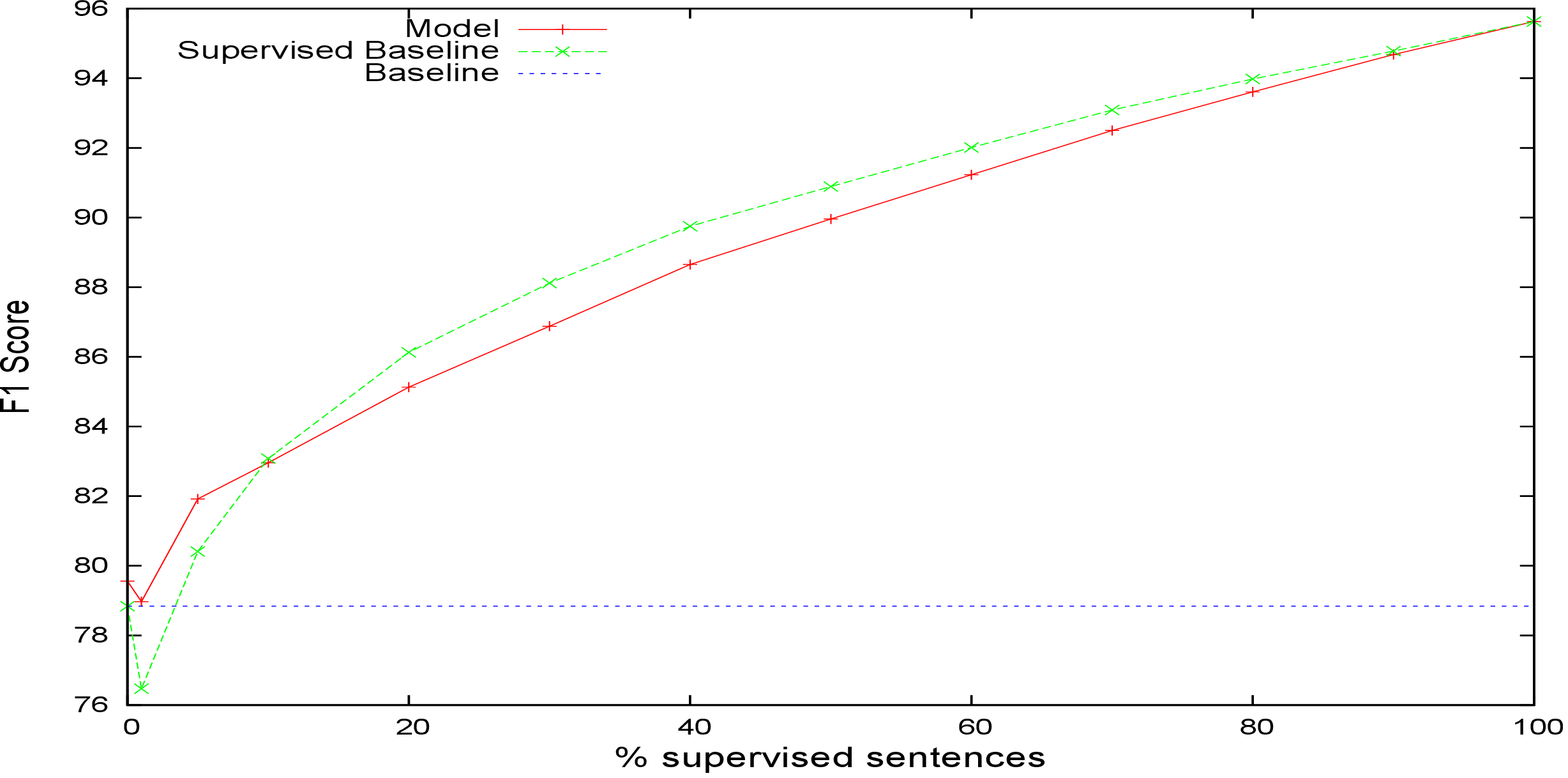}
    \vspace{-2mm}
    \caption{English}
    \label{fig:monolingual_semi_supervised_EN}
  \end{subfigure}
  \vspace{-3mm}
  \caption{F1 with a portion of the data labeled.}
  \vspace{-3mm}
\end{figure}

Figures \ref{fig:monolingual_semi_supervised_DE} and \ref{fig:monolingual_semi_supervised_EN} show the performance variation with \(S\). We make the following observations: 
\begin{itemize}[leftmargin=*]
\item In both languages, at around \(S=10\), the supervised baseline starts outperforming the semi-supervised model, which suggests that manually labeling about 10\% of the sentences is a good enough alternative to our training procedure. Note that 10\% amounts to about 3.6k sentences in German and 4k in English. We noticed that the proportion of seen predicates increases dramatically as we increase the proportion of supervised sentences. At 10\% supervised sentences, the model has already seen 63\% of predicates in German and 44\% in English. This explains to some extent why only 10\% labeled sentences are enough.
\item For German, it takes about 3.5\% or 1260 supervised sentences to have the same performance increase as 1.5M unlabeled sentences (Line 1 to Line 2 in Table \ref{tab:results}). Adding about 180 more supervised sentences also covers the benefit obtained by alignments in the multilingual model (Line 2 to Line 3 in Table \ref{tab:results}). There is no noticeable performance difference in English.
\end{itemize}
We also evaluated the performance variation on a completely unseen CoNLL test set. Since the test set is very small compared to the training set, the clustering evaluation is not as reliable. Nonetheless, we broadly obtained the same pattern.


\section{Related Work}
\label{sec:RelatedWork}
As discussed in section \ref{ssec:PreviousWork}, our work is closely related to the crosslingual unsupervised SRL work of \newcite{titovcrosslingual}. The idea of using \emph{superlingual} latent variables to capture cross-lingual information was proposed for POS tagging by \newcite{naseem2009multilingual}, which we use here for SRL. In a semi-supervised setting, \newcite{pado2009cross} used a graph based approach to transfer semantic role annotations from English to German. \newcite{furstenau2009graph} used a graph alignment method to measure the semantic and syntactic similarity between dependency tree arguments of known and unknown verbs.

For monolingual unsupervised SRL, \newcite{swier2004unsupervised} presented the first work on a domain-general corpus, the British National Corpus, using 54 verbs taken from VerbNet. \newcite{garg2012unsupervised} proposed a Bayesian model for this problem that we use here. \newcite{titov2012bayesian} also proposed a closely related Bayesian model. \newcite{grenager2006unsupervised} proposed a generative model but their parameter space consisted of all possible linkings of syntactic constituents and semantic roles, which made unsupervised learning difficult and a separate language-specific rule based method had to be used to constrain this space. Other proposed models include an iterative split-merge algorithm \cite{lang2011unsupervised} and a graph-partitioning based approach \cite{lang2011unsupervisedEMNLP}. \newcite{marquez2008semantic} provide a good overview of the supervised SRL systems.

\section{Conclusions}
\label{sec:Conclusions}

We propose a Bayesian model of semantic role induction (SRI) that uses crosslingual latent variables to capture role alignments in parallel corpora.  The crosslingual latent variables capture correlations between roles in different languages, and regularize the parameter estimates of the monolingual models.  Because this is a joint Bayesian model of multilingual SRI, we can apply the same model to a variety of training scenarios just by changing the inference procedure appropriately.  We evaluate monolingual SRI with a large unlabeled dataset, bilingual SRI with a parallel corpus, bilingual SRI with annotations available for the source language, and monolingual SRI with a small labeled dataset. Increasing the amount of monolingual unlabeled data significantly improves SRI in German but not in English. Adding word alignments in parallel sentences results in small, non significant improvements, even if there is some labeled data available in the source language.  This difficulty in showing the usefulness of parallel corpora for SRI may be due to the current assumptions about role alignments, which mean that only a small percentage of roles are aligned. Further analyses reveals that annotating small amounts of data can easily outperform the performance gains obtained by adding large unlabeled dataset as well as adding parallel corpora.

Future work includes training on different language pairs, on more than two languages, and with more inclusive models of role alignment.

\section*{Acknowledgments}
This  work  was  funded  by  the  Swiss  NSF  grant 200021\_125137 and EC FP7 grant PARLANCE.

\bibliographystyle{acl}
\bibliography{garg2016}

\end{document}